\documentclass{article}
\usepackage{graphicx}
\usepackage{booktabs}
\usepackage{amsmath}
\usepackage{geometry}
\usepackage{listings}
\usepackage{float}
\usepackage{listings}
\usepackage{xcolor}
\usepackage{url}
\usepackage{hyperref}

\lstdefinestyle{promptstyle}{
    backgroundcolor=\color{lightgray!20},
    basicstyle=\footnotesize\ttfamily,
    breaklines=true,
    frame=tb,
    framerule=0pt,
    captionpos=b,
    aboveskip=1em,
    belowskip=1em
}

\title{LLM-BI: Towards Fully Automated Bayesian Inference with Large Language Models}
\author{Yongchao Huang\footnote{Author email: yongchao.huang@abdn.ac.uk}}
\date{06 August 2025}

\begin{document}

\maketitle

\begin{abstract}
\noindent A significant barrier to the widespread adoption of Bayesian inference is the specification of prior distributions and likelihoods, which often requires specialized statistical expertise. This paper investigates the feasibility of using a Large Language Model (LLM) to automate this process. We introduce LLM-BI (Large Language Model-driven Bayesian Inference), a conceptual pipeline for automating Bayesian workflows. As a proof-of-concept, we present two experiments focused on Bayesian linear regression. In Experiment I, we demonstrate that an LLM can successfully elicit prior distributions from natural language. In Experiment II, we show that an LLM can specify the entire model structure, including both priors and the likelihood, from a single high-level problem description. Our results validate the potential of LLMs to automate key steps in Bayesian modeling, enabling the possibility of an automated inference pipeline for probabilistic programming.
\end{abstract}

\section{Introduction}

A significant barrier to the widespread adoption of Bayesian inference is the specification of prior distributions and likelihoods. While priors are powerful for incorporating domain knowledge, selecting and parameterizing them requires statistical expertise that many practitioners may not necessarily have. This can lead to the use of overly vague, "uninformative" priors or a reluctance to use Bayesian methods altogether \cite{ohagan2006uncertain}. This work investigates the feasibility of using a Large Language Model (LLM) to bridge this gap. We hypothesize that an LLM can act as an expert statistical consultant, translating a user's beliefs and problem description, expressed in natural language, into well-defined and appropriate components for a Bayesian model. This work serves as a foundational proof-of-concept for a fully automated Bayesian inference pipeline, which we term LLM-BI.

\section{Related Work}

This work contributes to a growing line of research exploring how Large Language Models (LLMs) can be integrated with statistical modeling and causal inference. Although the idea of leveraging LLMs to encode prior knowledge is becoming increasingly popular, current approaches vary widely in both their objectives and implementation strategies.

Some research uses LLMs to generate \textit{structural priors} rather than parameter priors. For example, Zhang et al. \cite{zhang_leveraging_2024} leverage an LLM to construct a similarity graph between concurrent causes (e.g. actors in a film). This graph serves as a structural prior to regularize a causal model, based on the principle that similar causes should have similar effects. Similarly, in machine learning contexts such as video retrieval, Jiang et al. \cite{jiang_prior_2024} use LLM encoders to refine feature embeddings, treating the semantic knowledge within the LLM as a "prior" to improve inter-concept relations. While these methods successfully incorporate high-level knowledge, they do not address the core Bayesian task of specifying tractable probability distributions for model parameters.

A second line of research, more aligned with our own, uses LLMs to directly elicit parameter distributions. In a study concurrent with ours, Capstick et al. \cite{capstick_autoelicit_2025} demonstrate that LLMs can provide expert prior distributions for predictive linear models, showing significant improvement over uninformative priors in low-data clinical settings. Their work also provides a valuable comparison between prior elicitation and in-context learning. Nafar et al. \cite{nafar_extracting_2025} focus on parameterizing Bayesian Networks (BNs), using an LLM to generate the conditional probability tables (CPTs) given a fixed network structure, effectively using the LLM as an expert to populate the model's parameters.

A third line of research focuses on understanding the LLM's own internal beliefs. Zhu and Griffiths \cite{zhu_eliciting_2024} present a compelling framework using iterated learning to elicit the implicit, human-like priors that an LLM holds about various phenomena. Their work treats the LLM as the subject of study, asking "What does the LLM believe?" This contrasts with our objective, which is to build a tool for a human user, asking "Can the LLM help a user formalize their beliefs?". Our work builds directly on the theoretical foundation laid out by Huang \cite{huang2025llmprior}, which introduced the concept of an \textit{LLM-Prior}: a formal operator that translates unstructured context into a valid probability distribution, and proposed a method for aggregating such priors.

The primary novelty of LLM-BI lies in its creation of a \textbf{PPL-free interface} for Bayesian modeling.
By leveraging an LLM as a universal translator between natural language and the domain-specific language of probabilistic programming languages (PPLs), our framework enables non-experts to construct sophisticated Bayesian models with ease. Our experiments demonstrate different levels of automation are possible, from specifying only the priors (Experiment I) to generating the entire probabilistic model, including the likelihood (Experiment II). A PPL engineer can provide this LLM interface, through which a user simply expresses their beliefs about the model and its parameters in natural language. The LLM-BI pipeline then translates these beliefs into structured distributions that can be directly consumed by any PPL backend, thereby significantly lowering the barrier to entry for applied Bayesian inference.
While the work of Capstick et al. \cite{capstick_autoelicit_2025} and Nafar et al. \cite{nafar_extracting_2025} successfully automates the parameterization of specific, pre-defined model classes (linear models and BNs, respectively), our Experiment II demonstrates a more general capability: generating the \textit{entire model structure}, including both priors and the likelihood function, from a single, holistic natural language description.

\section{Methods, Experiments and Results}
We examine the feasibility of building the LLM-BI pipeline via empirical verification. The general architecture of our pipeline consists of four main components: (1) a natural language interface for user input; (2) an LLM prompter that translates the input into a structured JSON object representing the statistical model; (3) a dynamic model builder that parses the JSON file, returned by the LLM, to construct a probabilistic model; and (4) an inference engine which performs inference (e.g. MCMC sampling) based on the specified probabilistic model.
We conducted two experiments using this pipeline with a simple yet fundamental model: Bayesian linear regression.
\begin{enumerate}
    \item \textbf{Experiment I (Partially Automated BI):} we test the LLM's ability to perform prior elicitation (as in \cite{huang2025llmprior}), a critical sub-task in Bayesian modeling.
    \item \textbf{Experiment II (Fully Automated BI):} we test the LLM's ability to perform end-to-end model specification, generating both priors and the likelihood from a holistic problem description.
\end{enumerate}

Both experiments used a synthetic dataset generated from a linear model $y = \alpha + \beta x + \epsilon$, with true parameters $\alpha=2.5$, $\beta=1.8$, and $\sigma=15.0$. All models were fitted using PyMC \cite{salvatier_probabilistic_2015}, and the LLM used was Google's \textit{Gemini} v2.5 model \cite{comanici2025gemini25pushingfrontier}.

\subsection{Experiment I: LLM-Elicited Priors}

This experiment compared two models. \textbf{Model 1} used standard, weakly informative priors for all parameters: $\alpha \sim \mathcal{N}(0, 100)$, $\beta \sim \mathcal{N}(0, 50)$, and $\sigma \sim \text{HalfNormal}(50)$.
\textbf{Model 2} used priors generated by the LLM from the following beliefs expressed in natural language:
\begin{itemize}
    \item \textbf{For $\alpha$}: "This is the intercept. I'm not very certain about it. I think it's probably around 0, but it could reasonably be as low as -25 or as high as 25."
    \item \textbf{For $\beta$}: "This is the slope. I strongly believe it's positive. My best guess is around 1.5 or 2. It's very unlikely to be greater than 10."
    \item \textbf{For $\sigma$}: "This is the model's error. It must be a positive number. Based on the data's spread, a value around 15 seems plausible."
\end{itemize}
For each parameter, the corresponding belief was inserted into the following LLM prompt:
\begin{lstlisting}[style=promptstyle, title={LLM Prompt for Experiment I}, label=lst:prompt1]
You are an expert statistician translating a user's belief into a PyMC prior.
Available distributions: "Normal", "HalfNormal", "Uniform", "Exponential".
Required JSON format: {"distribution": "Name", "params": {"param1": value1}}
USER BELIEF for '{parameter_name}': "{belief_text}"
Your response MUST be only the valid JSON object.
\end{lstlisting}

\subsubsection{Results}
\begin{figure}[H]
    \centering
    \includegraphics[width=0.9\textwidth]{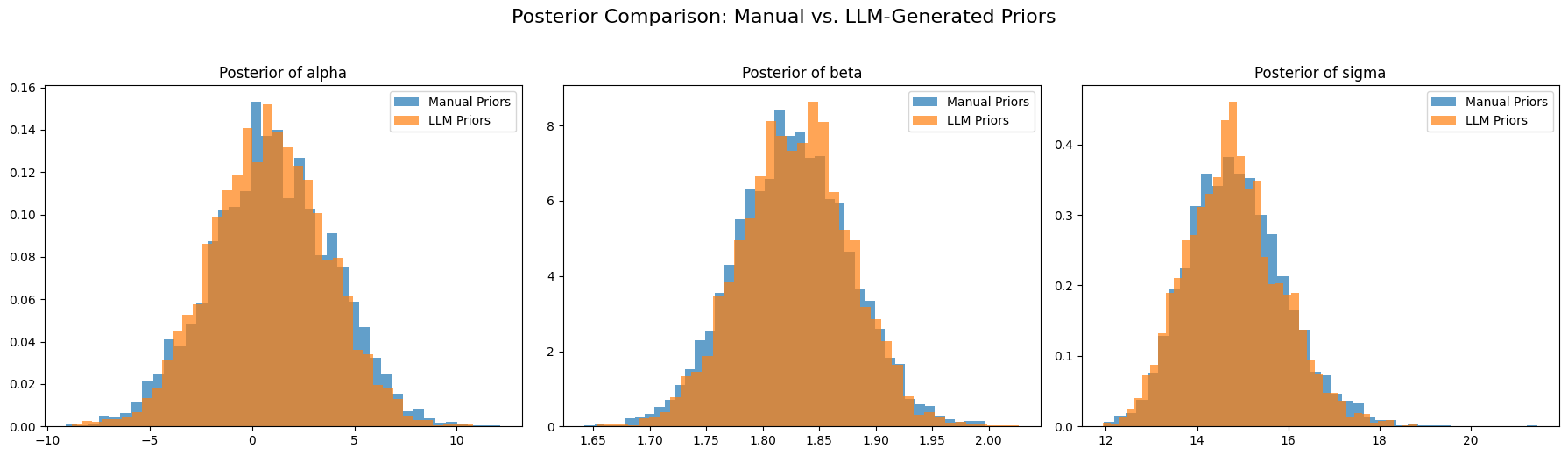}
    \caption{Exp. I: Comparison of posteriors from the Manual Priors and LLM Priors models.}
    \label{fig:exp1_comparison}
\end{figure}

The LLM successfully translated the beliefs into appropriate priors. The full specifications returned by the LLM in this run were\footnote{During our experiments, it is interesting to note that, when repeating this experiment with another call to the LLM, the LLM's choices could change, e.g. a previous run chose $\beta \sim \text{Exponential}(\lambda=0.5)$, which still appropriately captures the user's belief about the slope's likely value.}: $\alpha \sim \mathcal{N}(0, 12.5)$, $\beta \sim \mathcal{N}(2, 1)$, and $\sigma \sim \text{HalfNormal}(15)$. 
We then ran both the manually specified and LLM-specified models. The inferred results are shown in Fig.\ref{fig:exp1_comparison} and Table.\ref{tab:exp1_summary}.
The results clearly show that the posterior distributions from both models are nearly identical. This is visible in the overlapping histograms in Fig.\ref{fig:exp1_comparison} and confirmed by the summary statistics in Table.\ref{tab:exp1_summary}. While the means and modes are very close, it is notable that the standard deviation (`sd`) for every parameter in the LLM-specified model is slightly lower than in the manual model, suggesting a marginal increase in precision. Nonetheless, as expected, the strong signal from the 100 data points overwhelmed the subtle differences in the priors. This experiment successfully demonstrates that an LLM can act as a reliable "prior elicitation expert," correctly interpreting nuanced statements about uncertainty and shape to produce valid priors.

\begin{table}[H]
    \centering
    \caption{Exp. I: Numerical summary of posterior distributions.}
    \label{tab:exp1_summary}
    \begin{tabular}{lrrrrrrr}
        \toprule
        \multicolumn{8}{c}{\textbf{Model 1: Manual Priors}} \\
        \cmidrule(r){1-8}
        Parameter & mean & mode & sd & hdi\_3\% & hdi\_97\% & ess\_bulk & r\_hat \\
        \midrule
        alpha & 0.975 & 0.607 & 2.980 & -4.650 & 6.433 & 1660 & 1.0 \\
        beta & 1.827 & 1.823 & 0.051 & 1.734 & 1.924 & 1758 & 1.0 \\
        sigma & 14.881 & 14.661 & 1.078 & 12.928 & 16.906 & 2252 & 1.0 \\
        \bottomrule
        \\
        \toprule
        \multicolumn{8}{c}{\textbf{Model 2: LLM Priors}} \\
        \cmidrule(r){1-8}
        Parameter & mean & mode & sd & hdi\_3\% & hdi\_97\% & ess\_bulk & r\_hat \\
        \midrule
        alpha & 0.828 & 0.788 & 2.789 & -4.393 & 5.842 & 1974 & 1.0 \\
        beta & 1.829 & 1.843 & 0.048 & 1.744 & 1.923 & 2047 & 1.0 \\
        sigma & 14.805 & 14.742 & 1.032 & 12.906 & 16.716 & 2665 & 1.0 \\
        \bottomrule
    \end{tabular}
    \par
    \small\textit{Note: `hdi` stands for Highest Density Interval. `ess\_bulk` is the bulk effective sample size, and `r\_hat` is a convergence diagnostic.}
\end{table}

\subsection{Experiment II: Fully Automated Bayesian Inference}

We design a second experiment aiming to test if an LLM could automate the entire model specification from a single, high-level problem description. The objective was for the LLM to generate a complete model blueprint (priors and likelihood) in a single JSON object. The user's input was consolidated into the following narrative:
\begin{quote}
"I want to model the relationship between two variables, an independent variable 'X' and a dependent variable 'y'. I have a strong belief that the relationship is linear, so y should depend on X through an intercept and a slope. The relationship isn't perfect, so I expect there to be some normally distributed random error.

Here are my beliefs about the model parameters:\\
- The intercept, which we can call 'alpha', is probably around 0, but it could reasonably be anywhere between -25 and 25.\\
- The slope, let's call it 'beta', should definitely be a positive value. My best guess is that it's around 1.5 or 2.\\
- The error term, or noise standard deviation, which we can call 'sigma', must be a positive number. Based on a quick look at the data's spread, a value around 15 seems plausible."
\end{quote}
This description was embedded into the following, more complex prompt:
\begin{lstlisting}[style=promptstyle, title={LLM Prompt for Experiment II}, label=lst:prompt2]
You are an expert Bayesian statistician. Your task is to translate a user's problem description into a complete model specification in JSON format. The model should include priors for all parameters and a likelihood function.

The final JSON object MUST have two top-level keys: "priors" and "likelihood".

1. The "priors" key should map to an object where each key is a parameter name (e.g. "alpha") and the value specifies its distribution and parameters.
   - Available distributions: "Normal", "HalfNormal", "Uniform", "Exponential".

2. The "likelihood" key should map to an object with two keys:
   - "distribution": The name of the likelihood distribution (e.g. "Normal").
   - "formula": A string representing the mean of the likelihood (e.g. "alpha + beta * X"). The variables in this formula must correspond to the keys in the "priors" object and the data variable 'X'. The standard deviation of the likelihood should be the parameter named 'sigma' from the priors.

Here is the user's problem description:
---
{description}
---

Your response MUST be only the valid JSON object representing the complete model structure. Do not include any other text or markdown.
\end{lstlisting}

\subsubsection{Results}
The LLM successfully processed the holistic description and generated a complete and valid model structure. It correctly identified the linear relationship, formulated it as \texttt{"alpha + beta * X"}, and chose it as the mean for a Normal likelihood. The priors and likelihood it selected were:
\begin{itemize}
    \item Priors: $\alpha \sim \text{Uniform}(-25, 25)$, $\beta \sim \text{Exponential}(\text{rate}=0.5)$, and $\sigma \sim \text{HalfNormal}(\text{scale}=15)$.
    \item Likelihood: $y \sim \mathcal{N}(\mu = \alpha + \beta X, \sigma)$.
\end{itemize}
These choices were appropriate and consistent with the user's beliefs. Notably, the LLM's choice of an Exponential distribution for the slope \texttt{beta} correctly enforces positivity while having a mean of 2.0 (1/rate), aligning well with the user's "best guess."

\begin{figure}[H]
    \centering
    \includegraphics[width=0.9\textwidth]{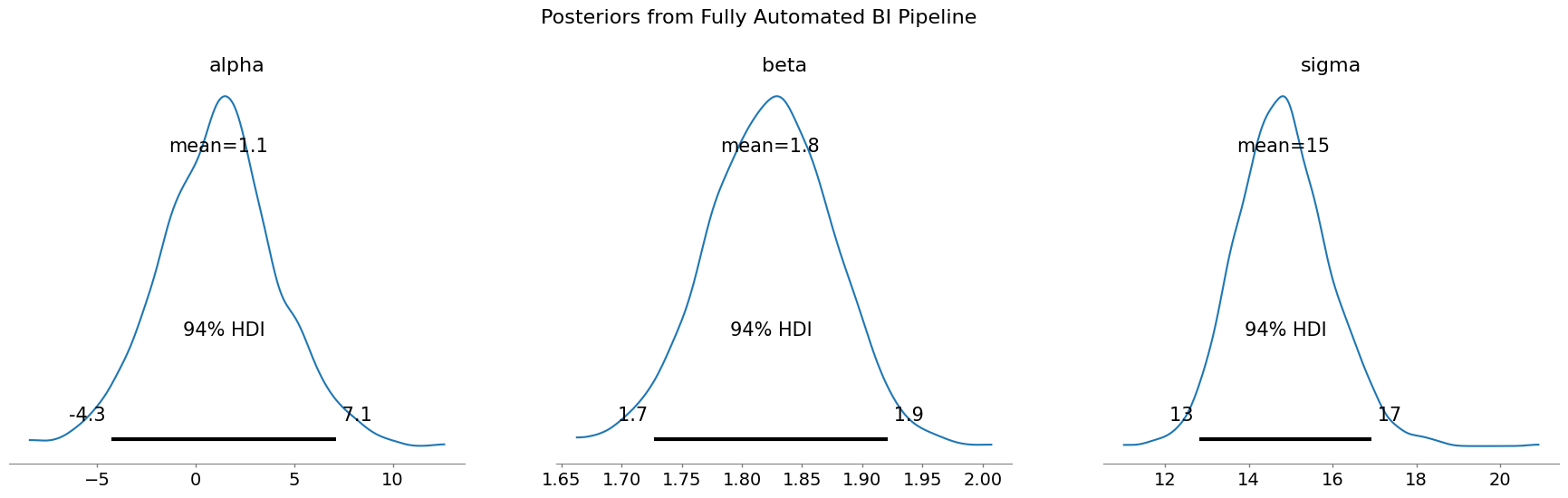}
    \caption{Exp. II: Posterior distributions from the fully automated LLM-BI pipeline.}
    \label{fig:exp2_posteriors}
\end{figure}

The posterior inference results from the fully automated model are shown in Fig.\ref{fig:exp2_posteriors} and Table.\ref{tab:exp2_summary}. We observe that, the model converged successfully (\texttt{r\_hat=1.0}) and accurately recovered the true data-generating parameters for the slope and noise level. For each parameter, the posterior mean, mode, and 94\% HDI are tightly centered around the true values (e.g. posterior mean for $\beta$ was 1.823 vs. ground truth value of 1.8). This result proves that the LLM is capable of acting as a "model architect," designing a statistically sound model from a high-level, text description, which can then be executed by an inference engine.

\begin{table}[H]
    \centering
    \caption{Exp. II: Numerical summary of posteriors from the fully automated model.}
    \label{tab:exp2_summary}
    \begin{tabular}{lrrrrrrr}
        \toprule
        Parameter & mean & mode & sd & hdi\_3\% & hdi\_97\% & ess\_bulk & r\_hat \\
        \midrule
        alpha & 1.146 & 1.484 & 2.970 & -4.264 & 7.131 & 1672 & 1.0 \\
        beta & 1.823 & 1.828 & 0.052 & 1.727 & 1.921 & 1617 & 1.0 \\
        sigma & 14.818 & 14.805 & 1.112 & 12.816 & 16.926 & 2220 & 1.0 \\
        \bottomrule
    \end{tabular}
\end{table}

\section{Conclusion}
This study successfully demonstrates the potential of Large Language Models to significantly lower the barrier to entry for Bayesian inference. Experiment I validated the LLM's role as a "prior elicitation expert", while Experiment II confirmed its more advanced capability as a "model architect" in a fully automated Bayesian inference pipeline.

This LLM-BI framework correctly interpreted user intent and designed statistically sound models. The resulting posteriors were not only consistent with those from a manually specified model but also accurately recovered some true parameters of the underlying data-generating process. While these experiments were conducted in a data-rich scenario where the likelihood dominated, the utility of such a system is expected to be even greater in sparse-data settings where prior specification is more critical. This work establishes a foundation for future research into a new paradigm of \textit{natural language-based probabilistic programming}, paving the way for more accessible and user-friendly tools for automated Bayesian data analysis.

\section*{Code Availability}
The code used in this work is available at: \url{https://github.com/YongchaoHuang/llm_bi}

\bibliography{references}
\bibliographystyle{plain}

\end{document}